\def\year{2021}\relax
\documentclass[letterpaper]{article} 

\usepackage{aaai21}  
\usepackage{times}  
\usepackage{helvet} 
\usepackage{courier}  
\usepackage[hyphens]{url}  
\usepackage{graphicx} 
\usepackage{natbib}  
\usepackage{caption} 
\usepackage{algorithm}
\usepackage{algorithmic}
\usepackage{amsmath}
\usepackage{amssymb}
\usepackage[switch]{lineno}  %
\usepackage{color}
\usepackage{booktabs}
\usepackage{threeparttable}
\usepackage{subfigure}
\usepackage{multirow}
\usepackage{enumitem}
\usepackage{bm}

\title{Series Saliency: Temporal Interpretation for Multivariate Time Series Forecasting}

\author{Qingyi Pan, Wenbo Hu, Jun Zhu\\}
\affiliations{
	Tsinghua University,\\
	\texttt{pqy19@mails.tsinghua.edu.cn, i@wbhu.net, dcszj@tsinghua.edu.cn}
	
}
\begin{document}

\maketitle

\begin{abstract}
Time series forecasting is an important yet challenging task. Though deep learning methods have recently been developed to give superior forecasting results, it is crucial to improve the interpretability of time series models. Previous interpretation methods, including the methods for general neural networks and attention-based methods, mainly consider the interpretation in the feature dimension while ignoring the crucial temporal dimension. In this paper, we present the series saliency framework for temporal interpretation for multivariate time series forecasting, which considers the forecasting interpretation in both feature and temporal dimensions. 
By extracting the ``series images'' from the sliding windows of the time series, we apply the saliency map segmentation following the smallest destroying region principle.
The series saliency framework can be employed to any well-defined deep learning models and works as a data augmentation to get more accurate forecasts. Experimental results on several real datasets demonstrate that our framework generates temporal interpretations for the time series forecasting task while produces accurate time series forecast. 
\end{abstract}

\section{Introduction} 
\noindent Time series is the data configured with a sequential order~\cite{hamilton1994time}, which is widely used in various applications with representative tasks as classification, forecast, imputation, etc~\cite{keogh2003need}.  
Traditional time series forecasting methods are often formulated as parametric models with a shallow architecture, such as the Box–Jenkins methods~\cite{box1976time} and the structural time series models~\cite{harvey1990forecasting}. 
Such methods adopt some explicit model assumptions that are easy-to-interpret, while their strict assumptions typically restrict the predictive capabilities and wide applications.

\begin{figure}[htbp] 
  \centering 
  \includegraphics[scale=0.265]{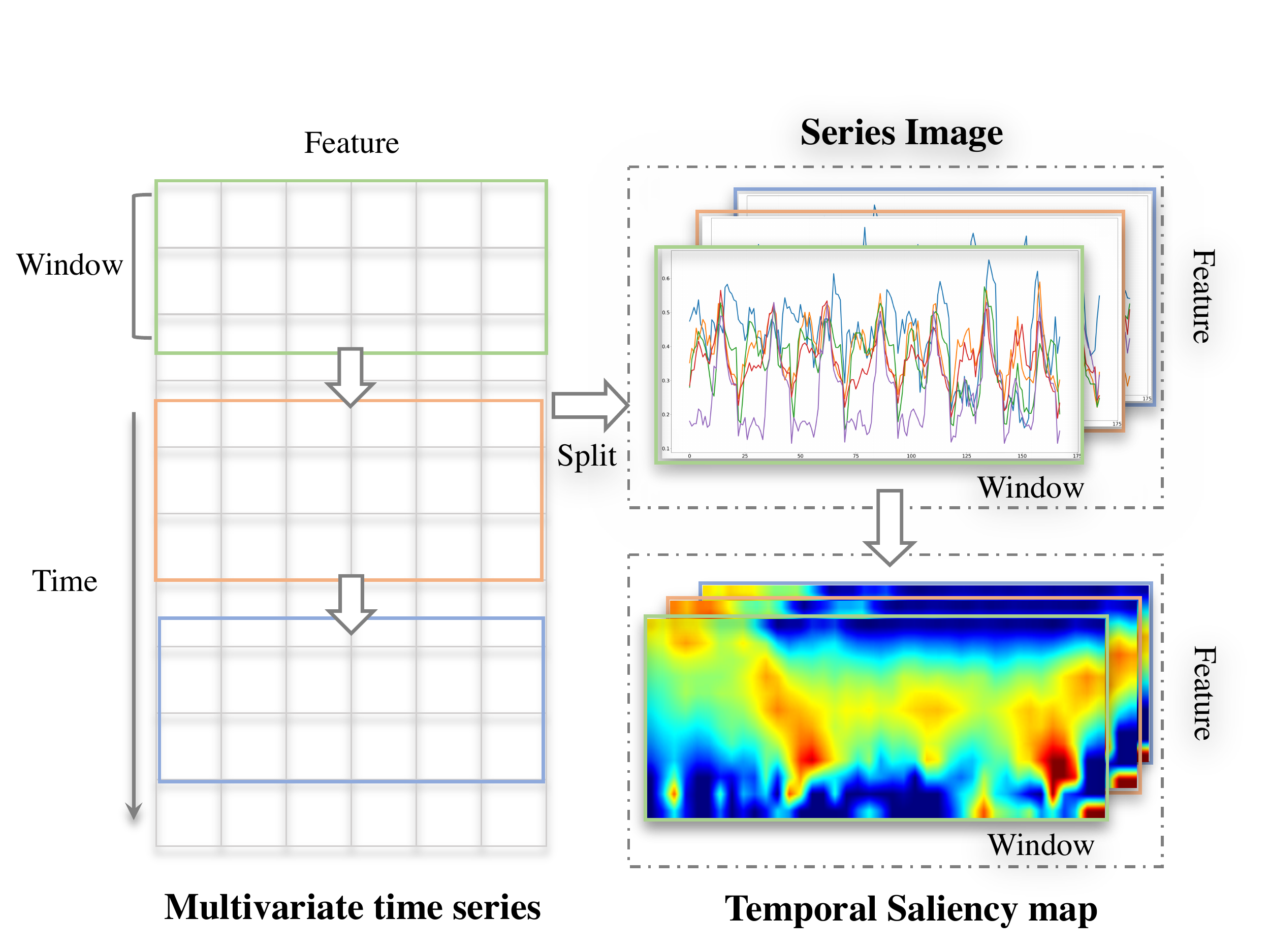}
  \caption{Multivariate time series and correspond temporal saliency map. The left is the original  multivariate time series data. The right is the generated  ``\textbf{series image}" $ \in \mathbb{R}^{w \times D}$  by sliding the window along the time axises, where $w$ represents the window size and $D$ represents the feature dimension. For each ``series image" corresponding to a temporal saliency map, finding which parts of features are most important to forecasting.}
  \label{fig:MultiTime} 
\end{figure}

More recently, deep learning methods have become increasingly effective and popular for time series forecasting. The representative deep models for time series forecasting~\cite{gamboa2017deep} include recurrent neural networks~(RNN), long-short term memory~(LSTM)~\cite{hochreiter1997long}, gated recurrent unit~(GRU)~\cite{chung2014empirical} and neural attention methods. 
Although effective, deep learning methods are typically treated as a black-box and difficult to interpret the outputs~\cite{castelvecchi2016can}, which stands as a crucial bottleneck of the further applications of such models. In particular for the time series forecasting task, the model interpretations should explain not only the single-point output but also the temporal trends and changes. 

A straightforward solution is to use the interpretation methods for general neural networks, such as LIME~\cite{ribeiro2016should}, DeepLift ~\cite{shrikumar2017learning} and Shap ~\cite{lundberg2017unified}. These methods use gradient information to extract feature information for single-time forecasts after the back-propagation training. However, these methods do not consider the temporal information for forecasting interpretation, which leads to single-point explanations for temporal sequential forecasts. 
Another line of research on time series forecast interpretations is to migrate the attention methods from the sequence-based language or speech representation methods~\cite{bahdanau2014neural,vaswani2017attention}. The attention mechanism uses sequence align functions for sequence recurrent neural networks and is found effective for sequence predictions tasks, such as language modeling and machine translation. 
However, the attention values for explaining RNNs are calculated via the related importance of the different time steps and there are doubts that they are based on the intermediate feature importance instead of model interpretations~\cite{serrano2019attention,jain2019attention}.

In this work, we jointly consider both the feature and temporal information and propose the series saliency interpretable framework, which is a temporal interpretable multivariate time series forecast model with saliency maps. As shown in Figure~\ref{fig:MultiTime}, we consider the temporal multivariate time series as a $\textit{window} \times \textit{feature}$ ``series image" and use the series saliency framework to segment the most meaningful and useful part for the time series forecast task. Following the smallest destroying region~(SDR) principle~\cite{dabkowski2017real, chang2017interpreting}, we use the masking technique to get heatmaps of feature importance for one input series image.
With the masking techniques, the added Gaussian noise or blur can also improve the robustness and overall forecasting performance, as shown in \cite{devries2017improved}. 
We present both quantitative and qualitative results on several typical time series datasets, which show that our  method provides temporal interpretations for the time series forecasts and meanwhile achieves better or comparable forecast results.

\section{Preliminaries}
In this section, we introduce the useful preliminaries, including: 1) the time series forecasting task and interpretations and 2) the saliency map method used in computer vision and the effective smallest destroying region principle.

\subsection{Time Series Forecasting and Interpretations}
We consider interpreting multivariate time series forecasting results. Formally, we use $X$ to denote a series of observed time series signals, where every single signal $x_t$ is a real vector with dimension $D$:
\begin{equation}
    X=[x_1,x_2,\cdots, x_t, \cdots]^{\top},~~ x_{t}\in \mathbb{R}^{D}.
\end{equation}

For every time step $t$, we aim to predict the future time value after a given horizon $\tau$: $x_{t+\tau}$.
In most of the cases, the horizon of the forecasting task is chosen according to the task settings. For example, for the traffic usage, the horizon $\tau$ of interest ranges from an hour to a day; for the stock market data, even second or minute ahead forecast can be meaningful for generating returns. 

After getting the forecasting results, we are interested in the question of the forecasting interpretation: \textit{what features most and least contribute to the final forecast predictions?}
The traditional statistical methods, such as Box-Jenkins methods~\cite{box1976time} and structural methods~\cite{harvey1990forecasting}, are easy to interpret because they adopt the explicit model assumptions that can extract interpretations directly from the learned model parameters. 
Though deep learning methods have superior prediction capabilities~\cite{heaton2018ian}, they are hard to interpret since the deep model assumptions are stacked with multiple non-linear activations or blocks. 
Some works tried to use the explainability methods for general neural networks to obtain the time series forecasting feature importance, such as LIME~\cite{ribeiro2016should}, DeepLift ~\cite{ribeiro2016should} and Shap ~\cite{lundberg2017unified}. However, these methods do not capture the temporal information for forecasting. Attention mechanism calculates the related importance values among different time steps, so it is hard to capture the interdependence among features well at the same time~\cite{serrano2019attention, jain2019attention}. 
As for interpreting the time series forecasts, feature importance aligns through the temporal dimension, which is the key challenge in our paper. Our method jointly consider the feature and temporal information via the saliency map method and can produce more global and applicative forecasting interpretations. 

\subsection{Saliency Map and Masks for Smallest Destroying Region}

Saliency map~\cite{kadir2001saliency} is a widely-used method in computer vision, which segments the parts of inputs that are important for a model's outputs. One popular principle to learn saliency maps is the smallest destroying region~(SDR)~\cite{dabkowski2017real}, which learns the smallest informative removing region of the input that preserves a confident output. 

We denote an input image as $X$ with a corresponding class $c$. By adding noise (e.g., Gaussian blur) to the input image, we get the reference image $\hat{X}$. 
Following the SDR principle, we aim
to learn a mask $M$ and part of a image will be wiped out by applying the mask to it.
The mask $M$ is assigned to each pixel $x_i$ from a image and the value of the mask will be between $0$ and $1$: $m(x_i) \rightarrow [0, 1]$. Here $m(x_{i})=0$ means that the original pixel is used and $m(x_{i}) = 1$ means that the reference region added some noise is used. 
Through the definition of the mask, we can introduce a local perturbation to the image as:
\begin{equation}
\label{eqn:mask_image}
 \widetilde{X} = M \odot \hat{X}+(1-M) \odot X,  
\end{equation}
where $\odot$ is the element wise dot product.
The SDR saliency map prevents the confident classification with the smallest destroying region~\cite{dabkowski2017real}:
\begin{equation}
    \label{eqn:sdr}
    \min_{M} P_{g}(c|M \odot \hat{X}+(1-M) \odot X) + \lambda \mathbb I(M),
\end{equation}
where $P_{g}(c|X)$ is the output probability of the class c,  $g(.)$  is the classification model, $\lambda$ is the regularization parameter and $\mathbb I(M)$ is a complexity regularization function of the mask $M$. The obtained mask represents the feature importance of the input.

Dabkowski and Gal~\cite{dabkowski2017real} amortizes the cost of computes saliency map by auxiliary neural networks. Fong~\cite{fong2017interpretable,fong2019understanding} solves for input feature saliency map as the parameters of the mask that optimize the SDR objective. 
In this paper, we formulate the time series as ``series images'', as shown in Figure~\ref{fig:MultiTime}, and propose the series saliency method to interpret the temporal multivariate time series forecast. 

\begin{figure}[htbp] 
  \centering 
  \includegraphics[scale=0.27]{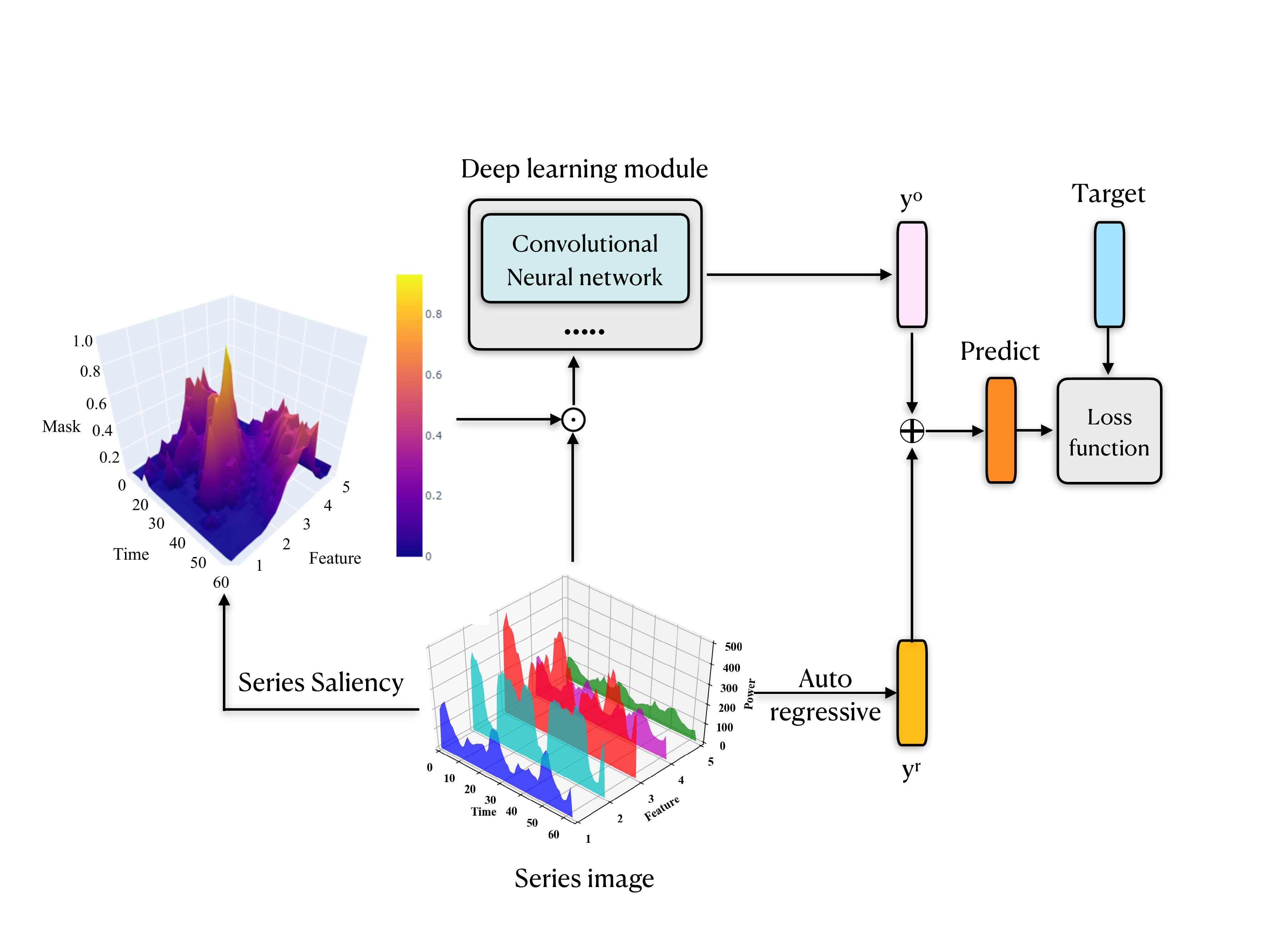}
  \caption{Series Saliency composed of series saliency module, deep learning module and an linear auto-regressive module.}
  \label{Fig_Archi} 
\end{figure}

\section{Series Saliency Method}
We introduce the series saliency method: the windows of time series is formulated as ``series images'', then the saliency method is proposed to interpret the time series forecasting task and finally  result is presented as interpretation that outputs the feature importance with temporal information.  

\subsection{Series Image and Mask}
As shown in Figure~\ref{fig:MultiTime}, multiple data of different time windows are obtained from the multivariate time series through the time line, where each row of the time series corresponds to one feature dimension in the multivariate time series. 
$w$ is the length of window and $D$ is the feature dimensions. We define ``series images'' $X$ for the data of every sliding window with the size $\textit{window} \times \textit{feature}$. 
By adding noise (constant, noise or blut) to the input series image $X$, we have three ways to obtain the reference ``series image'' $\hat{X}$, with reference elements $\hat{x}_{t,i}$ defined as:
\begin{equation}
\label{eqn:xref}
\hat{x}_{t,i}=
\left\{
\begin{array}{rcl}
x_{t,i} && \rm constant \\
x_{t,i} + \epsilon_{\sigma_{1}} && \rm noise \\
g_{\sigma_2}(x_{t,i}) && \rm blur \\
\end{array},
\right.
\end{equation}{}
where $\epsilon_{\sigma_{1}} \sim \mathcal{N}(\mu,\,\sigma_{1}^{2})\ $ is a Gaussian noise and $g_{\sigma_2}$ is a Gaussian blur kernel on feature $x_{t,i}$ with the maximum isotropic standard deviation $\sigma_2$. In order to find a mask $M$ assigning to each feature $x_{t,i}$, the value of mask will be between $0$ and $1$: $m(x_{t,i})  \rightarrow [0,1]$. 
Here $m(x_{t,i})=1$ means that the original feature is replaced by the noisy feature in reference $\hat{X}$, and $m(x_{t,i})=0$ means the original value is reserved. 
Through the mask, we use Equation~\ref{eqn:mask_image} to get the series images $X$ and regard their references $\hat{X}$ as the perturbations for the original time series. 

\subsection{Model Architecture and Training}
The series saliency for forecasting model contains three components: 1) the series saliency module, 2) a deep learning module and 3) a linear auto-regressive module. 
The series saliency uses masks to add perturbations on the original time series. The masked ``series image'' will be processed by a well-designed deep neural network module to extract representations and obtain time series forecasts $y_o$. 
One merit of our series saliency method is that any deep learning method can be used here, such as convolution neural network~\cite{lecun1989generalization}, recurrent neural network~\cite{elman1990finding}, Long short-term memory(LSTM)~\cite{hochreiter1997long}, long- and short-term temporal pattern neural network(LSTNet)~\cite{lai2018modeling}, transformer-based self-attention encoder~\cite{vaswani2017attention} and etc.
Then, we used a linear autoregressive model to learn the linear patterns for multivariate time series. Denote the forecasting result of the AR component as $y^r$, and the coefficient of the autoregressive model as $W_{k} \in \mathbb{R}^{D}$ and $b \in \mathbb{R}$, where $p$ represents the order of autoregressive model AR($p$). The AR model is formulated as:
\begin{equation}
    y^{r} = \sum_{k=0}^p W_{k}^{\top} y_{t-k} + b.
\label{eq_ar}
\end{equation}{}

The final forecasting $\hat{y}$ of our proposed method is then obtained by intergrating the outputs of the non-linear and linear component:
\begin{equation}
    \hat{y} = y^{r} + y^{o}.
\end{equation}

When training the series saliency model, we follow the empirical loss minimization while limiting the complexity of Mask, i.e., $\mathbb{I}(M)$ in Equation~\ref{eqn:sdr}.
In the series saliency method, we consider two parts for the mask complexity $\mathbb{I}(M)$: 1) the size of masks should be as small as possible and 2) the masks shapes should be smooth.

We denote $\ell_{p} = \phi(x_{t+\tau}, f_{c}(X, M))$ as the multivariate time series forecasting prediction part. We design $\phi(.)$ to measure model performance, where $\phi(.)$ represents the mean squared error of $(x_{t+\tau}$ and $f_{c}(X, M))$, or other metrics.
The penalty function for the mask size is defined as $\ell_{m}$: 
\begin{equation}
    \ell_{m}=\|M\|_{p_0},
\end{equation}
where the matrix norm $\|M\|_{p_0}$ is set as 2-norm or 3-norm. 
Another penalty function for the mask smoothness is defined as $\ell_{r}$:
\begin{equation}
\label{eqn:mask_smooth}
\ell_{r} = \sum_{(t,i)}(m_{t,i}-m_{t,i+1})^2 + \sum_{(t,i)}(m_{t,i}-m_{t+1,i})^2.
\end{equation}
In training procedure, noise injection or blur on masks region is actually a data-augmentation method~\cite{wen2020time}.
Our model can learn most salient data characteristics by data augmentation and  achieve better performance. The strategies can help network focuses on the most salient data characteristics \cite{tsipras2018there,cubuk2019randaugment, cubuk2018autoaugment}.

Overall, the training minimizing objective $L_1$ can be:
\begin{equation}
  L_1 =  \ell_{p} + \lambda_{1} \ell_{m} + \lambda_{2} \ell_{r},
\end{equation}
where $\lambda_1$ and $\lambda_2$ are two regularization parameters.
We use gradient descent to optimize the objective function $L_1$. In each step of training iterations, we sample mini-batch of $n$ samples series image $X^{(b)}$, correspond horizon $\tau$ of $x_{(t+\tau)}^{(b)}$, reference series images $\hat{X}^{(b)}$, mask $M$(in previous definition). Firstly, we use Equation~\ref{eqn:mask_image} to generate perturbations $\widetilde{X}^{(b)}$ for original series image. The $\widetilde{X}^{(b)}$ is feed into deep learning module and linear auto-regressive module separately. $y^o$ obtained by deep learning module and $y^r$ obtained by the auto-regressive module is integrated together to got $\hat{y}$. Then the final prediction part $\ell_{p}$, and regularization  term ($\ell_{m}$ $\ell_{r}$) are calculated simultaneously to got objective function $L_{1}$. Then the parameters of deep learning module, linear module and mask are updated simultaneously by the back-propagation.

\subsection{Forecast Interpretation}
After training the forecasting model by optimizing $L_{1}$, we use a subsequent procedure to learn forecast interpretations through the smallest destroying region (SDR) principle~\cite{dabkowski2017real}, as defined in Equation~\ref{eqn:sdr}.
Given an ``series image'' $x_0$, this interpretation procedure is to summarize  compactly the effect of deleting feature regions, either setting values to zeros or Gaussian noise, in order to explain the behaviour of the deep neural net forecastors. The SDR principle is finding the higher salient feature regions by identifying the highly representative mask. 
The SDR interpretation procedure considers a deletion game and  the goal is to find the smallest deletion mask $m$ for input $X$ that regression have the worst performance, i.e.,to maximize the forecast error $\ell_{p}$.
The loss function of the interpretation procedure is:
\begin{equation}
    L_{2} = - \ell_{p} + \lambda_{1} \ell_{m} + \lambda_{2} \ell_{r},
\end{equation}
where $\ell_m$ and $\ell_r$ are also used to limit the complexity of the masks. 
In this procedure, we fix the learned model parameters of the forecasting modules including the deep learning model and the AR model. By the above definition, we sample a series image $X^{(s)}$ in the test set, mask $M$ and reference series image $\hat{X}^{(s)}$. 
$\widetilde{X}^{(s)}$ is feed into series  for forecasting to obtain $L_2$ loss. Then only the parameter of mask can be updated by back-propagation in interpretable process.

The overall training and interpretation process is outline in Algorithm~\ref{alg:process}. In algorithm,  the number of training steps of mask component  $k_2$ is the hyperparameter.  The mask parameter is $\theta_{k}$. 

\begin{algorithm}
    \begin{algorithmic}[1]
    \FOR{number of training iterations}
    \STATE Sample mini-batch of $n$ samples series image $x^{(b)}$ and correspond horizon of $x^{(b)}_{t+\tau}$ from training dataset; $M$ represents the mask.
    \STATE Update the series saliency parameter including mask component by descending its stochastic gradient:
    \STATE $\bigtriangledown_{\theta} \frac{1}{n} \sum_{i=1}^{n} ( (x^{b}_{t+\tau}, f_{\theta}(x,M)) + \lambda_{1}\| 1-M \|_{1} + \lambda_{2} \ell_{r} )$;
    \ENDFOR
    \STATE Get accuracy on test set, and select one unseen samples ($x^{(s)},y^{(s)})$) in test set. 
    \FOR{$k_2$ steps}
    \STATE Only update mask parameters $\theta_{k}$ to obtain series saliency map by descending its stochastic gradient:
    \STATE $\bigtriangledown_{\theta_{k}} (-\ell_{p}    + \lambda_{1} \ell_{m} + \lambda_{2} \ell_{r})$;
    \ENDFOR
    
\end{algorithmic}
    \caption{Minibatch stochastic gradient descent training of series saliency.} 
    \label{alg:process}
\end{algorithm}

\subsection{Feature exchangeability}
As shown in Figure 1, after the series saliency, we can obtain the series saliency map in the interpretable procedure, where each row corresponds to the dynamic changes of feature importance in time series. Due to the mask smoothness penalty $\sum_{(t,i)} (m_{t,i}-m_{t,i+1})^2$ between features in Equation \ref{eqn:mask_smooth}, it assumes that there exists some unknown correlation between features.  

For some multivariate time series datasets, like the electricity dataset used in the paper, the feature dimension means the powerplant position and nearby features correspond to the nearby stations. Thus the order of feature and time dimension of ``series image'' is both fixed. In this case, one can use our series saliency framework to obtain interpretations directly. While for other series data contains exchangeable features (like the air quality or the industry stock indicies), we may modify our algorithm slightly. In interpretation procedure, we remove the $\sum_{(t,i)}(m_{t,i}-m_{t,i+1})^2$ term in smoothness penalty, and use $l_r^{'}$ instead,
\begin{equation}
    l_r^{'}=\sum_{(t,i)}(m_{t,i}-m_{t+1,i})^2   
\end{equation}{}
which means $l_r^{'}$ only consider the correlation through the time line. After the interpretation procedure, the temporal feature importance of each feature is calculated. We use simulated annealing to search the permutation. It makes sure that the nearby feature have the correlative importance to the forecasting results. 

Given an ``series image'' $X^{(s)}$, we can obtain the dynamic feature importance in $M^{(s)}$ during the interpretation procedure. We could formulate it as travelling salesman problem. Label the features with the numbers $\{f_1...f_D\}$ and define:

\begin{equation}
   x_{i,j}= \left\{
   \begin{aligned}
    1 & & { i \rightarrow  j} \\
    0 & & \rm{ other }\\
    \end{aligned}
   \right.
\end{equation}

For $i=\{1...n\}$, take $d_{i,j}$ to the similarity between dynamic importance of feature $a$ and feature $b$ on permutation of mask $M$.
\begin{equation}
    d(f_a,f_b)=\sqrt{\sum_{i=1}^{T}|f_{(a,i)}-f_{(b,j)}|^2}
\label{Eqn:dist}
\end{equation}

Then the objective function can be written as the following problem:
\begin{equation}
    g(.) = {\rm min} \sum_{i=1}^{D} \sum_{j \neq i, j=1}^{D}( d(f_i, f_j) x_{i,j} )
\label{Eqn:obj}
\end{equation}

where $x_{i,j} \in \{0,1\}$. We use simulated annealing algorithm to solve the permutation. The simulated annealing consists of the following operators: initialization, neighbor selection, evaluation and output, where $\psi$ represents the temperature, $s$ is the current permutation of features, $v$ is the candidate solution, $g(.)$ is the objective value of current state $v$, $h(s)$ is to swap the order of two features in permutation.

\begin{algorithm}
\begin{algorithmic}[1]
\STATE Create the initial permutation of feature $s=\{f_1,f_2...f_D\}$
\STATE Input distance matrix $dis(s) \in R^{D \times D}$ calculated by $d(f_a,f_b)=\sqrt{\sum_{i=1}^{T}|f_{a,i}-f_{b,i}|^2}$
\STATE Set the initial temperature $\psi$
\WHILE{$\psi > \psi_0$}
\STATE Perturbing the current state $s$ to generate $v=h(s)$
\STATE Evaluate the new state $g(v)$ by Equation \ref{Eqn:obj}.
\IF{$g(v)$ satisfies the probabilistic acceptance criterion}
\STATE Update current state $s=v$
\ENDIF
\STATE Update $\psi$ according to the annealing schedule.
\ENDWHILE
\STATE Output current permutation $s$
\end{algorithmic}
\caption{Simulated Annealing for feature permutation} 
\label{alg:process}
\end{algorithm}

\section{Experiment and analysis}

For the empirical experiments, firstly we compare the performance of four widely-used deep learning models with and without the proposed series saliency method on three representative multivariate time series forecasting datasets. 
We also give an ablation study on the mask and AR model components. 
Then we show extensive qualitative results of how saliency heatmaps interpret the forecasts.

\subsection{Experimental Setting}
We test different methods on three representative time series forecasting dataset: electricity, air-quality and industry data. In Appendix~\ref{appendix:datasets}, we summarize the statistics of experiment datasets. More details about the datasets is given in Appendix~\ref{appendix:datasets}. For time series forecasting performance we use two evaluation metrics, relative squared error~(RSE) and empirical correlation coefficient~(CORR), which are defined in Appendix. For RSE, lower values are better. For CORR, higher values are better. Other experimental setups are given in Appendix.

\subsection{Deep Learning Modules}
\label{sec:methods}
We use four state-of-art deep learning methods for comparison(CNN, GRU+Attention, LSTNet and Self Attention encoder), more details are given in Appendix \ref{appendix:datasets}. These four deep learning models achieve superior forecasting performances and in the following results they are used as a deep learning module to get interpreted by the proposed series saliency method.

\begin{table*}[tp]
  \centering
  \fontsize{6.5}{8}\selectfont
  \begin{threeparttable}
    \begin{tabular}{ccccccccccc}
    \toprule
    \multicolumn{2}{c}{Dataset} & \multicolumn{3}{c}{Industry} &\multicolumn{3}{c}{Air Quality} &  \multicolumn{3}{c}{Electricity}  \cr
    \cmidrule(lr){1-2} \cmidrule(lr){3-5} \cmidrule(lr){6-8} \cmidrule(lr){9-11} 
     \multicolumn{2}{c}{} & \multicolumn{3}{c}{Horizon} &\multicolumn{3}{c}{Horizon} &  \multicolumn{3}{c}{Horizon}\cr
     \midrule
    Methods & Metrics & 3 & 6 & 12  & 3 & 6 & 12 & 3 & 6 & 12 \cr
    
    \midrule
    CNN & RSE & 0.1622 & 0.1682 & 0.1983   &0.3088&0.3620 & 0.4048 & 0.1004 & 0.1049 &0.1066  \cr
    without saliency & CORR &0.8968&0.9196&0.8107 &0.7749&0.7009&	0.6361&0.8829&0.8715&0.8661  \cr
     \midrule
     GRU  & RSE &0.2115&0.1913&0.2219 &0.3121&0.3574&0.3972 & 0.1192&0.1249&0.1316\cr
     without saliency  & CORR &0.9531&0.9359&0.9036& 0.8037& 0.7120&	0.6386&0.8789&0.8774&0.8671\cr
    \midrule
    LSTNet         & RSE& 0.1806&0.1901&0.2284  & 0.3268 & 0.3600 & 0.4072 & 0.0894 & 0.0923 & 0.0991 \cr
    without saliency  & CORR &0.9501&0.9335&0.8645 &0.7771&0.6775&0.6237 & 0.9264&	0.9157&0.9095 \cr
    \midrule
    Self-Attention & RSE  &0.1474&0.1518&0.1857 & 0.3007&0.3526&0.3871&0.0882&0.0921&0.1005 \cr
      without saliency  & CORR &0.9612&0.9418&0.9048 &\bf{0.8129}& 0.7223 &0.6396 &0.9193&0.9065 &0.9097\cr
    \midrule[1pt]
    
    CNN  & RSE & 0.1531 & 0.1762 & 0.1951 & 0.3071 & 0.3507 & 0.4022 & 0.0991 & 0.0993 & 0.1003\cr
     with saliency(Ours) & CORR & \bf{0.9617} & 0.9274 & 0.8816 & 0.7461 & 0.7283 & 0.6382 & 0.8991 & 0.8943 & 0.8951 \cr
    \midrule     
    GRU   &  RSE & 0.1771 & 0.1974 & 0.1996 & 0.2901 & \bf{0.3228} & \bf{0.3418}  & 0.1077 & 0.1116 &0.1121 \cr
     with saliency(Ours) & CORR & 0.9551 & 0.9255 & 0.9117 & 0.8095 & 0.7125 & 0.6492 & 0.9053 & 0.8875 & 0.8774\cr
    \midrule
    LSTNet & RSE & 0.1777 & 0.1791 & 0.2261 & 0.2993 & 0.3495 & 0.3869 & 0.0881 & 0.0942 & \bf{0.0936} \cr
     with saliency(Ours) &  CORR & 0.9504 & 0.9343 & 0.9011 & 0.7935 & \bf{0.7293} & \bf{0.6501} & 0.9284 & \bf{0.9173} & 0.9057 \cr
    \midrule
    Self-Attention  & RSE & \bf{0.1195} & \bf{0.1304} & \bf{0.1735} & \bf{0.2879} & 0.3505 & 0.3509 & \bf{0.0877} & \bf{0.0908} & 0.0985\cr
      with saliency(Ours) & CORR & 0.9553 & \bf{0.9591} & \bf{0.9165} & 0.8119 & 0.7093 & 0.6379 & \bf{0.9293} & 0.9125 & \bf{0.9106}
 \cr
    \bottomrule
    \end{tabular}
    \end{threeparttable}
     \caption{Results summary(in RSE, and CORR) of implemented methods on three multivariate time series datasets. Each row has the results of a specific method in a particular metric. Each column compares the results of all methods on a particular dataset with a specific horizon value. Bold face indicates the best result of each column in a particular metirc.}
  \label{tab:results}
\end{table*}

\subsection{Forecasting Results}

In Table \ref{tab:results}, we summarize the forecasting results of the four methods on the three datasets in two metrics.
We compare the four deep learning methods with and without the series saliency method. 
We set horizons as ${3,6,12}$, respectively. For example, the horizons was set from 3 to 12 hours for the electricity forecasting. The best result for each (data, metric) pair is highlighted in bold font.
By comparing the upper and lower groups of results in the table, for most settings with the same model and dataset, adding the series saliency would bring better forecasting results, i.e., lower RSE or larger CORR. This is because that the added series saliency masks work as a data augmentation method and strengthen the generalization for the testing data.
The performance improvements from the incorporated series saliency module are found more obvious on datasets with higher dimensions (the electricity with 321 dimensions VS air quality with 12 dimensions). 
For all eight comparing methods, the self-attention encoder with the series saliency obtains most of the best forecasting result and this is due to the powerful representation capability of the transformer encoder model.

\begin{figure}
    \centering
    \includegraphics[scale=0.13]{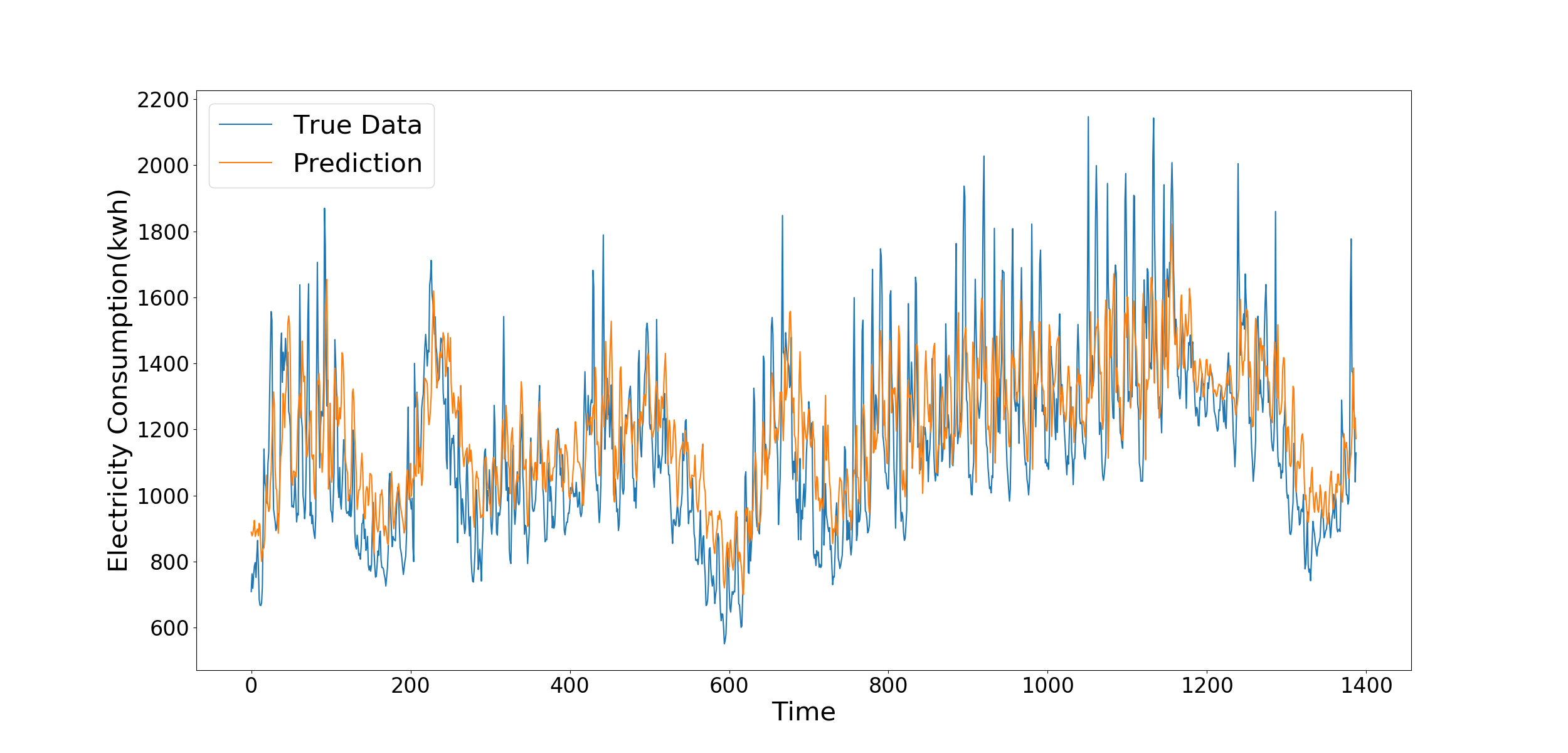}
    \caption{By using the self-attention encoder method with series saliency, Forecasting result of air quality for horizon=6. The model learn trends of orginal data.}
    \label{fig:forecast}
\end{figure}

Figure~\ref{fig:forecast} gives an example of forecasting results of the self-attention encoder with the series saliency on the electricity dataset. As can be seen, our method gives accurate forecasts that predicts multiple peaks and trends. 
The forecasting results on more datasets are deferred to Appendix ~\ref{appendix:forecast}.

\subsection{Ablation study}
 
\begin{figure}
    \centering
    \includegraphics[scale=0.6]{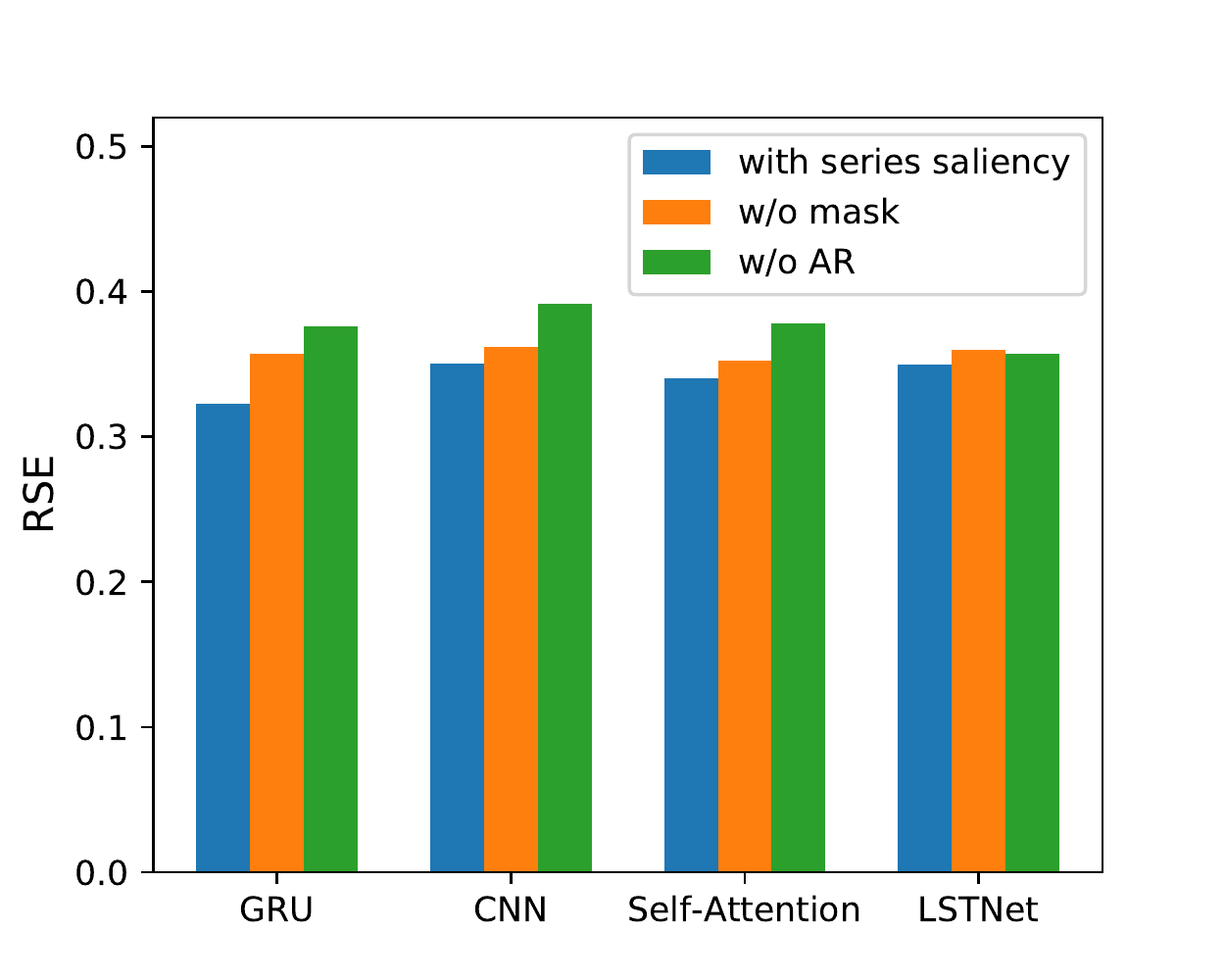}
    \caption{Ablation study of various model on horizon = 6 air quality. The results show that both Mask and auto-regressive component are  essential for forecasting.}
    \label{fig:augument}
\end{figure}

In order to test that the effectiveness of all model modules, we conduct an ablation study: delete every mask component and compare the forecasting results through the two metrics,  RSE and CORR. We name the different tested model with four deep learning modules in the ablation study as follows:
\begin{itemize}
\item \textbf{With series saliency} Four different deep learning models with series saliency, including mask and AR components
\item \textbf{Series saliency w/o mask} The series saliency model without mask component
\item \textbf{Series saliency w/o AR} The series saliency model without auto-regressive component.
\end{itemize}
In Figure~\ref{fig:augument}, we show the RSE results of the three different model settings with four deep learning modules on the air quality dataset with the horizon as $6$. 

As can be seen from the results, the masks and the introduced series saliency improve the forecasting result of four deep learning modules. Because the adding noise or Gaussian blur works as a data augmentation method.

All models with the AR module get lower RSE errors and this is because the AR module explicitly models the temporal and trend information.

\subsection{Qualitative Study on Interpretation Features by Series Saliency}
We give a qualitative study on the interpretation features of the series saliency method by analyzing the learned saliency mask heatmaps.

\begin{figure}[htbp] 
  \centering 
  \includegraphics[scale=0.42]{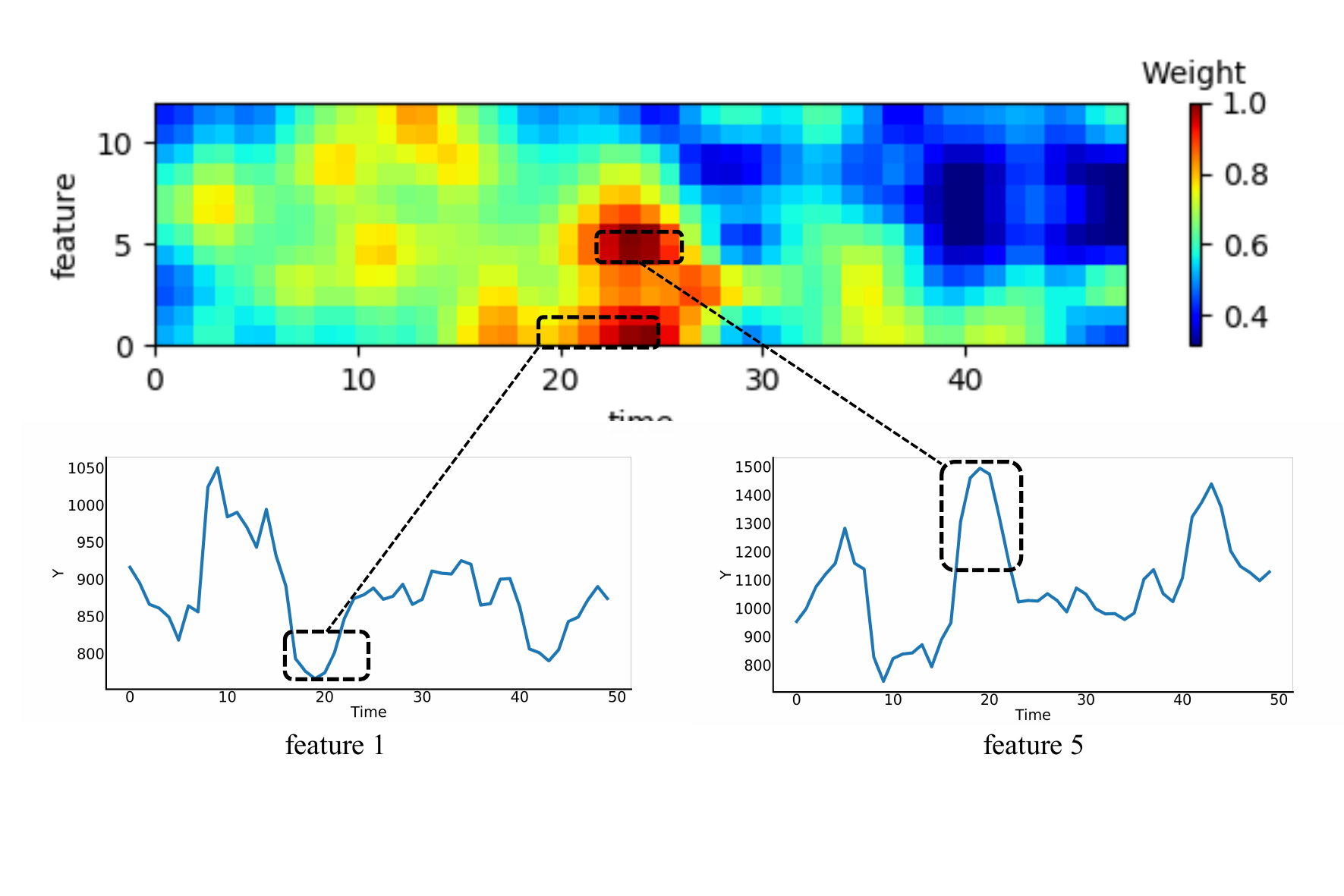}
  \caption{Correspond interpretable region in temporal saliency map. The series saliency is visualized on horizon=6 air quality data time series forecasting. The figure is best viewed in color.}
  \label{fig:heatmap_air} 
\end{figure}

\textbf{Visualization of Saliency Heatmap}
We apply the series saliency methods with the self-attention encoder module on the air quality dataset. 
In Figure~\ref{fig:heatmap_air}, we visualize the learned mask component when forecasting the the randomly-selected future value after the horizon $\tau=6$.
As can be seen from the saliency mask heatmap, the features in the time from $20$ to $30$ have the the largest saliency and this corresponds to the extreme low value in dimension $1$ which means the concentration of CO. Afterwards in the time from $30$ to $50$, the relatively low saliency corresponds to the more stationary concentration of CO. 
More saliency results on other datasets can be found in Appendix~\ref{appendix:feaure}.

\textbf{Correspondence between Saliency and Frequency}
In this experiment, we analyze the correspondence between saliency map region in the mask, the original time series data and the data frequencies. 
We run the self attention encoder method with series saliency method on the electricity method and show the results in Figure~\ref{fig:frequency}.
As can be seen from the results, the data of the channel 250 has the high value of the saliency mask and reflects a periodic structure (top left corner figure). 
To the contrary, the data of the channel 36 has the low value of the saliency mask and reflects a relatively acylic structure (top right corner figure).
In order to prove the periodicity of our selected channels, we map the corresponding feature to frequency domain by fast Fourier transform(FFT)\cite{nussbaumer1981fast} and show the  the frequency result below the data of the two channels.

\begin{figure}[htbp] 
  \centering 
  \includegraphics[scale=0.35]{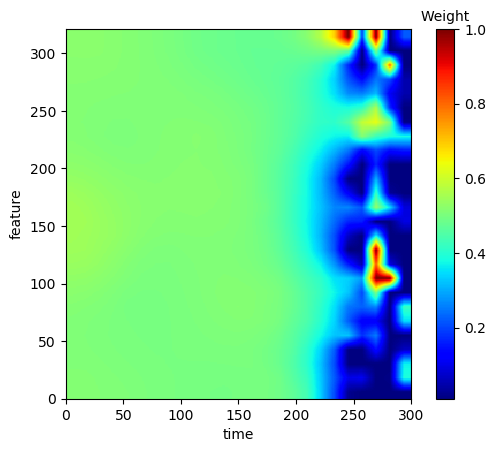}
  \caption{Series saliency for $AR(50)$ on  the electricity dataset with the horizon as $5$. As can be seen, within the time order 50, $AR(50)$ have distinct feature saliency while beyond the time order 50, AR(50) has no feaure saliency. The figure is best viewed in color.}
  \label{fig:ar_saliency} 
\end{figure}

\begin{figure*}[htbp] 
  \centering 
  \includegraphics[scale=0.4]{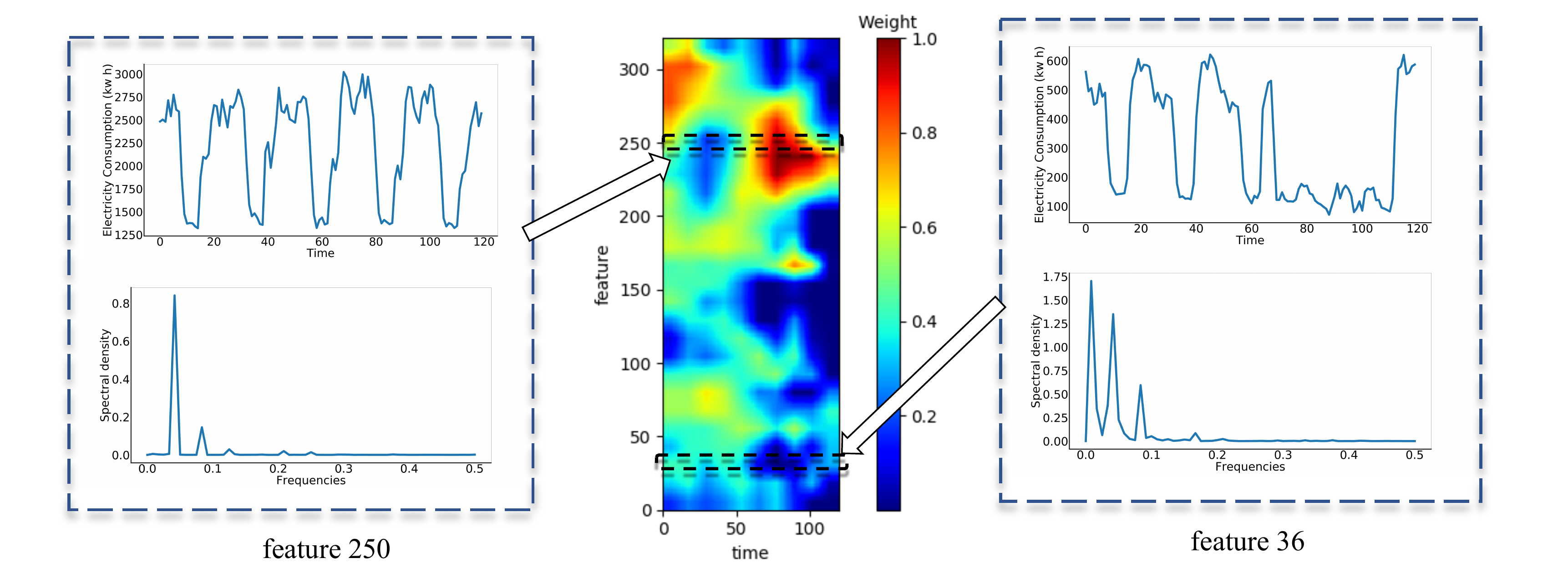}
  \caption{Select relevant information for settings on $horizon=6$ electricity with window size=120. The feature 250 corresponding to higher salient region is often periodic while the feature 36 corresponding to weaker salient region is acyclic. The figure is best viewed in color}
  \label{fig:frequency} 
\end{figure*}

\textbf{Series Saliency for AR Method, A Validation}
In order to validate the effective information learnt by mask, we design an validation experiment. We delete the deep learning module and only reserve 50-order autoregressive model to forecast for electricity on horizon=6. 
The experiment results is shown in Figure \ref{fig:ar_saliency}. 
The learned saliency maps within the time order is distinctly different from the maps beyond the time order. Within the time order, i.e., the time between 250 and 300, the saliency maps is concentrated on some specific feature which reflects that the AR(50) have specific feature saliency within the time order.
However when beyond the time order, i.e., the time between 0 and 250, the saliency mask has a saliency uniformity which reflects that AR(50) has no feature saliency beyond the time order.

\section{Related Works and Discussions}
In this section, we give related works and give discussions.

\subsection{Deep Learning for Time Series Forecast}
Deep learning model have been succssfully used for time series forecasting, like(CNN, RNN, or Transformer etc). To adapt convolutional neural network to time series datasets, multiple layers of convolutions filters were used to to capture past information for forecasting~\cite{borovykh2017conditional, oord2016wavenet}.
Also, LSTM or GRU \cite{hochreiter1997long} were developed to model long-range dependencies. \citet{lai2018modeling} proposed LSTNet, which is a model using a combination of CNN for short-term information and RNN for long-term information. 
Recently, Transformer is used as a new architecture which leverages the self-attention mechanism to process a sequence of data~\cite{parikh2016decomposable, vaswani2017attention}. 
\citet{li2019enhancing} tackles time series forecasting using transformer. It enhances the locality by using casual convolutions for each cell in each layer.
\citet{lim2019temporal} introduces the Temporal Fusion Transformer(TFT)-a novel attention-based architecture which combines the high-performance multi-horizon forecasting with considering temporal dynamic.

\subsection{Interpretation Methods for General NNs}
Deep learning methods for time series forecast yield accurate predictions, but it is insufficient to characterize the feature importance for model interpretability. Explaining methods for general DNNs can be used for obtaining the time series feature importance. LIME~\cite{ribeiro2016should} interprets model predictions based on locally approximating the model around a given prediction. However, the surrogate  samples used for interpretation are considered independently for different time steps. DeepLift~\cite{shrikumar2017learning} calculates the feature importance by selecting reference values, and calculating the changes of input to reference. But it does not consider the order of time steps, and features are considered independently for neighboring time steps. 
To interpret time series forecasting, we need to consider both the feature and temporal information and capture the temporal feature importance in multivariate time series.
Compared with the aforementioned methods, our method is able to capture the most meaning temporal feature information from the time series forecasts. 

\subsection{Saliency Map for Computer Vision}
Saliency map shows which parts of an image feature are important for network’s output, which is generally used in the computer vision applications. Examples include visualizing the middle layer of neural network \cite{simonyan2013deep}, mapping a specified class to a region in an image\cite{zeiler2014visualizing} and simulating learning procedure \cite{mahendran2015understanding} and so on. Specifically, masks~\cite{dabkowski2017real} propose two principles for learning saliency maps: 1) smallest destroying region: \citet{fong2017interpretable} proposes mask to “deleting” different regions of input feature for reinterpret network saliency map, which leading to more meaningful perturbation and hence explanations; and 2) smallest sufficient region: \citet{fong2019understanding} proposes to constrain the area of mask to a fixed value that maximize the model’s output, and use mask to interpret saliency map.

Recently, there are several works on the saliency map method for analyzing time series.
\cite{wang2018multilevel} proposed mWDN for time series classification to decomposed sub-series in different frequencies as inputss. mWDN exhibits the importance spectrum in saliency map and varying importance of all the features. But it does not capture the map between series data and importance spectrum. Compared with mWDN, our proposed method can generate saliency maps that give interpretations for time series forecasts and this type of interpretation gives both temporal and feature dimension information.

\cite{ren2019time} combines the spectral residual(SR) and the convolution neural network in time-series anomaly detection. They use SR model from visual saliency map to detect anomaly, and use CNN to improve performance. 
\subsection{Feature Exchangeability in Series Saliency}
As shown in Figure~\ref{fig:MultiTime}, the series image in the feature dimension is sensitive to the feature exchangeability due to the mask smoothness penalty in Equation~\ref{eqn:mask_smooth}.
Thus, a pre-defined feature ordering is recommended for our series saliency method. 
Also for some multivariate time series datasets, like the electricity dataset used in this paper, the feature dimension means the powerplant position and nearby features correspond to the nearby stations. In this case, one can use series saliency to obtain forecast interpretations directly. Otherwise, a clustering preprocess is recommended to make sure nearby feature have the correlative importance to the forecasting results.

\section{Conclusion}
In the paper, we presented a novel temporal interpretation framework (Series Saliency) for multivariate time series forecasting. By jointly consider both the feature and temporal information, we  model temporal feature importance successfully for the final predictions. Also the series saliency method can be employed to any deep learning models and works as data augmentation to improve performance. With in-depth analysis and series saliency visualization, we shows that our series saliency framework can do improve multivariate time series forecasting, while produces some meaningful interpretation by series heatmap.

For future work, there are several promising directions in extending the work. The deep learning model likes a neural basis in frequency domain. How to model the basis of deep learning model is an challenging model.

\bibliography{ref}

\clearpage
\appendix
\onecolumn

\section{Dataset Details}
\label{appendix:datasets}
\paragraph{Electricity}
The UCI electricity\cite{electrictyUCI} load diagrams data set contains 370 customer power consumption per unit time. There is no missing value in this data set, recording the power consumption per 15 minutes (KWH) from 2011 to 2014. Each column of data represents a customer (370 columns in total), each row represents a quarter (140256 rows in total), and all time labels are subject to Portuguese time. 

\paragraph{Air-Quality}
The dataset \cite{AirqualityUCI} contains 9358 instances of hourly averaged responses from an array of 5 metal chemical sensors embedded in Air quality Device. Data was recoreded from March 2004 to February 2005(one year). The device was located on the field in a significantly polluted area.

\paragraph{Industry data}
The data of Hangseng Stock Composite Index(HSCI) and another eleven industry stock indices are obtained from Wind platform. Eleven industry stock indicies include consumer good manufacturing, consumer service, energy, finance, industry, information technology, integrated industry, raw material, real estate, utilities. The dataset covers time period from September 2006 up to September 2019.

\begin{table}[htbp]
 \centering
 \begin{tabular}{lccl}
  \toprule
  Datasets & T & D & L \\
  \midrule
  Electricity    & 26304 & 321 &  1 hour\\
  Air-Quality    & 9358  &  12 & 1 hour \\
  Industry Stock Composite Index  & 3205 &  12 & 1 day\\
  \bottomrule
 \end{tabular}
 \caption{Dataset statistics, where $T$ is length of time series or data size, $D$ is the number of variables, $L$ is the sample rate.}
   \label{table:data}
\end{table}

\section{Metric Description}
\label{appendix:metrics}
On typical multivariate time series datasets, we followed the evaluate metrics \textbf{RSE} and \textbf{CORR} to measure the accuracy of models' forecasting ability. The first metric is the root relative squared error(\textbf{RSE}). The \textbf{RSE} are the scaled version of the widely used Root Mean Square Error(RMSE), which is designed to make more readable evaluation, regardless the data scale.

\begin{equation}
    RSE = \frac{\sqrt{\sum_{t=t_{0}}^{t_1} \sum_{i=1}^{n}(y_{t,i}-\hat{y_{t,i}})^2} }{\sqrt{\sum_{t=t_{0}}^{t_1} \sum_{i=1}^{n}(\hat{y}_{t,i}-\overline{\hat{y}_{t_0:t_1, 1:n}})^2 }}
\end{equation}{}

The second metric is the empirical correlation coefficient(\textbf{CORR}).

\begin{equation}
CORR = \frac{1}{n}\sum_{i=1}^{n} \frac{(y_{t,i} - \overline{\hat{y_{t_0:t_{1,i}}}})(\hat{y}_{t,i}-\overline{\hat{y}_{t_0:t_{1,i}}})}{\sqrt{ \sum_{t=t_0}^{t_1}(y_{t,i}-\overline{y_{t_0:t_{1:i}}})^2\sum_{t=t_0}^{t_1}{(\hat{y}_{t,i}-\overline{\hat{y}_{t_0:t_{1:i}}})^2}}}
\end{equation}

where $y$ and $\hat{y}$ is the ground truth and the predicted value. $\overline{Y}$ denotes the mean of set y. For \textbf{RSE}, the lower the better, whereas for \textbf{CORR}, the higher the better.

\section{Detailed Experimental Settings}

\subsection{Base module}

The four state-of-art deep learning methods for comparison(CNN,GRU,LSTNet and Self Attention encoder). We introduce more details of models.

\begin{itemize}
\item \textsl{CNN}: Convolutional nerual network \cite{heaton2018ian} designed to ensure only past information for forecasting. We use 7-layer CNN to do time series forecasting.
\item \textsl{GRU + Attention}: GRU \cite{chung2014empirical} had been used in time series forecasting and combined with attention to improve interpretability.
\item \textsl{LSTNet}: LSTNet \cite{lai2018modeling} is a model using CNN and RNN for multivariate time series forecasting. The architecture use convolutional network and the Recurrent Neural Network(RNN) to extract short-term local dependency patterns among variables and to discover long-term patterns for time series trends.
\item \textsl{Self Attention encoder}: Transformer-based \cite{li2019enhancing} architecture has been modified for forecasting. We design a variant of Transformer with encoder-only structure which consists of $L$ blocks of multi-head self-attention layers and position-wise feed forward layers.
\end{itemize}

\subsection{Hyperparameters}
Since the model structure is universal for all methods, we adjust the same optimal hyperparameters on the training data. Firstly, we use the Adam~\cite{kingma2014adam} algorithm for the optimization with learning rate $10^{-4}$ and weigth decay $10^{-3}$. For data preprocessing, we scale the data into the range $[0,1]$ by batch normalization \cite{ioffe2015batch} to avoid extreme values and improve the computation stability. The matrix norm of mask is set $p_0=2 ,3$. Also, we set the $\lambda_1=10^{-3}$ and $\lambda_2=10^{-3}$ in $L_1$ and $L_2$. We select the batch size correspond the size of dataset.

\section{Forecasting Visualization and Interpretation Features on More Datasets}
\label{appendix:forecast}

\begin{figure*}[htbp] 
  \centering 
  \includegraphics[scale=0.25]{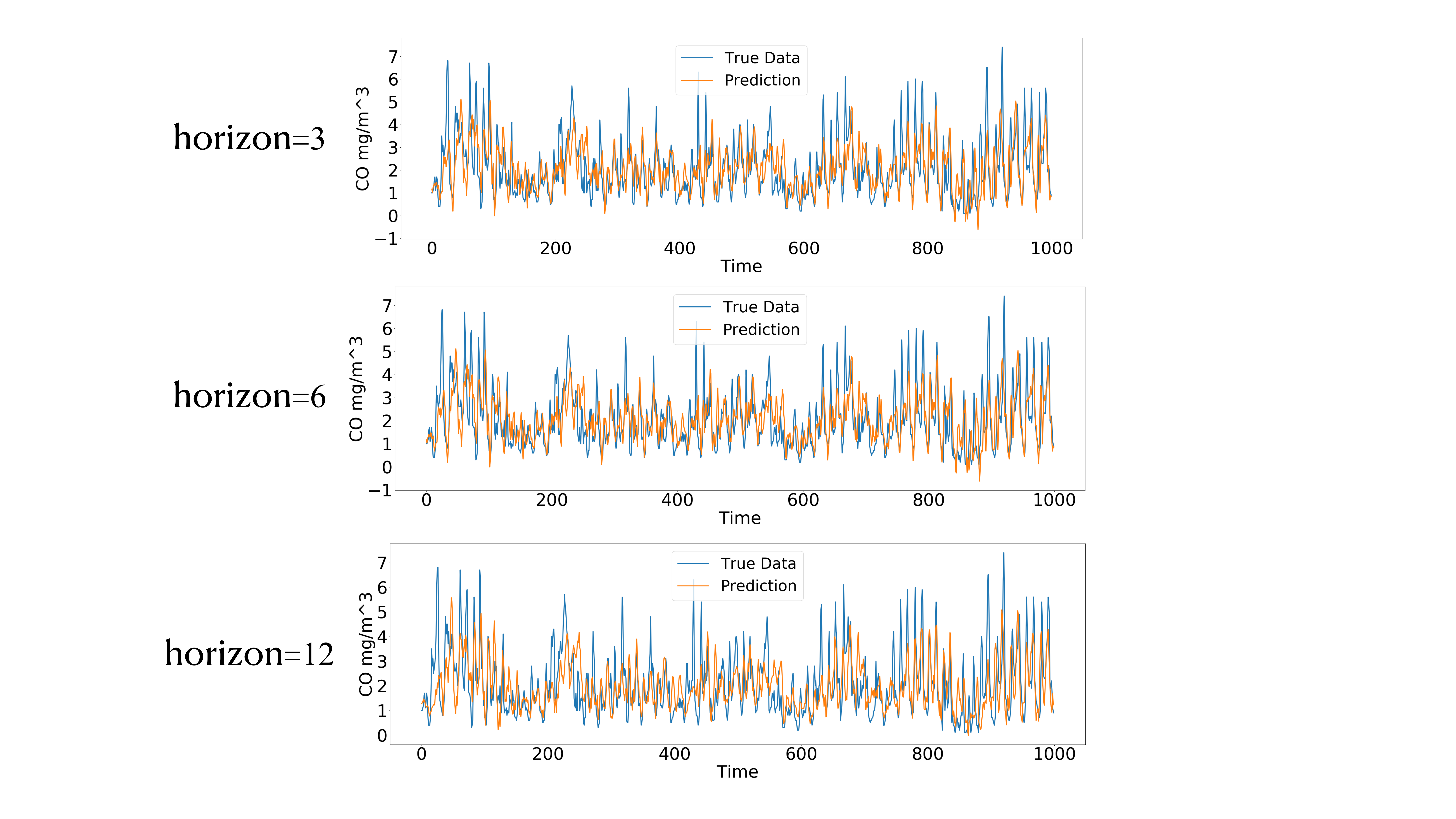}
  \caption{Prediction results for self-attention encoder in series saliency on air quality with  horizon=$3,6,12$ and window size=64. The feature is the true hourly averaged concentration CO in $mg/m^3$}
  \label{fig:air} 
\end{figure*}

\begin{figure*}[htbp] 
  \centering 
  \includegraphics[scale=0.27]{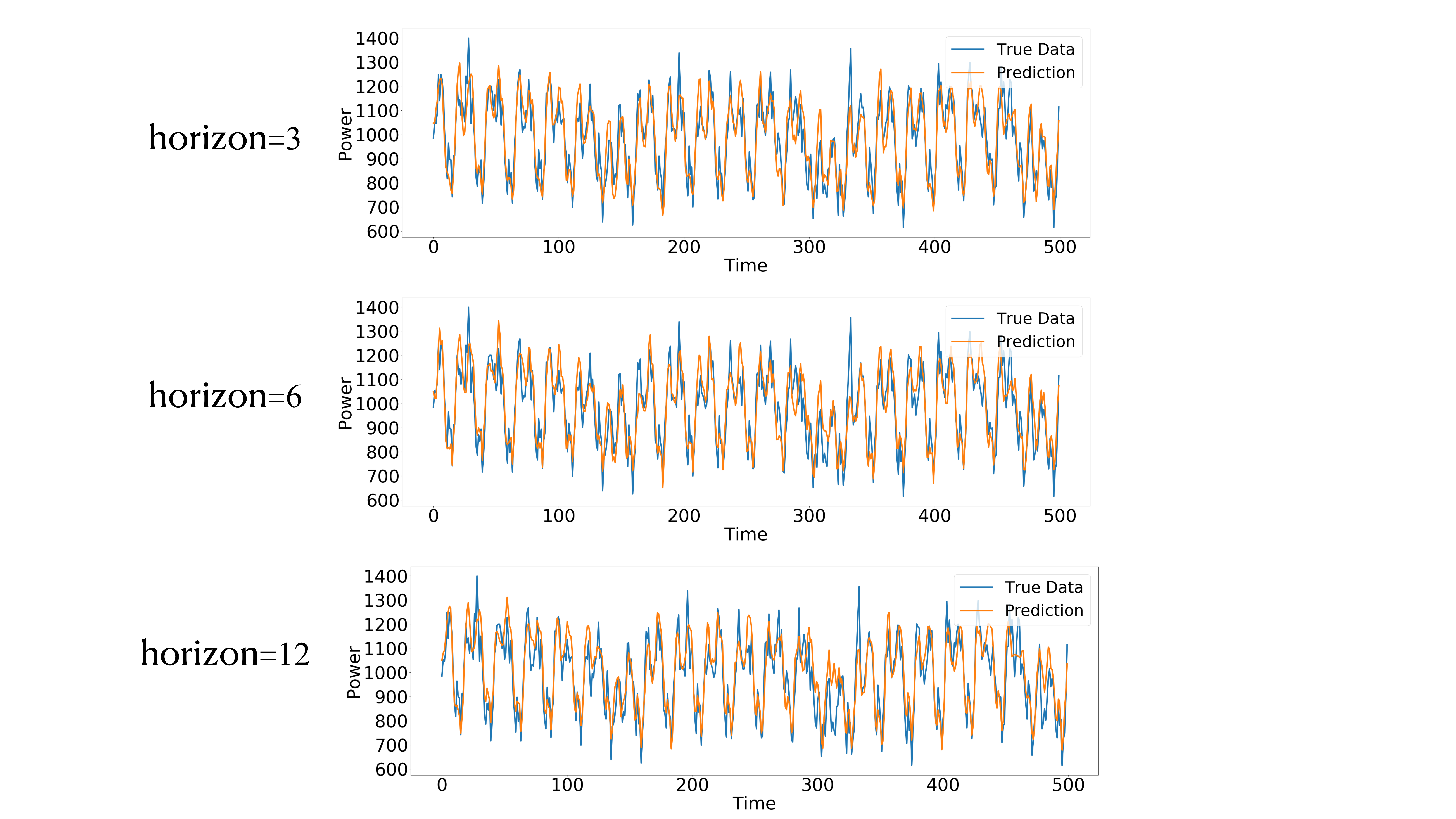}
  \caption{Prediction results for self-attention encoder in series saliency on electricity with  horizon=$3,6,12$  and window size=168. The feature is power consumption of No.7 powerplant. The model learned the periodicity of electricity data.}
  \label{fig:ele} 
\end{figure*}

\begin{figure*}[htbp] 
  \centering 
  \includegraphics[scale=0.26]{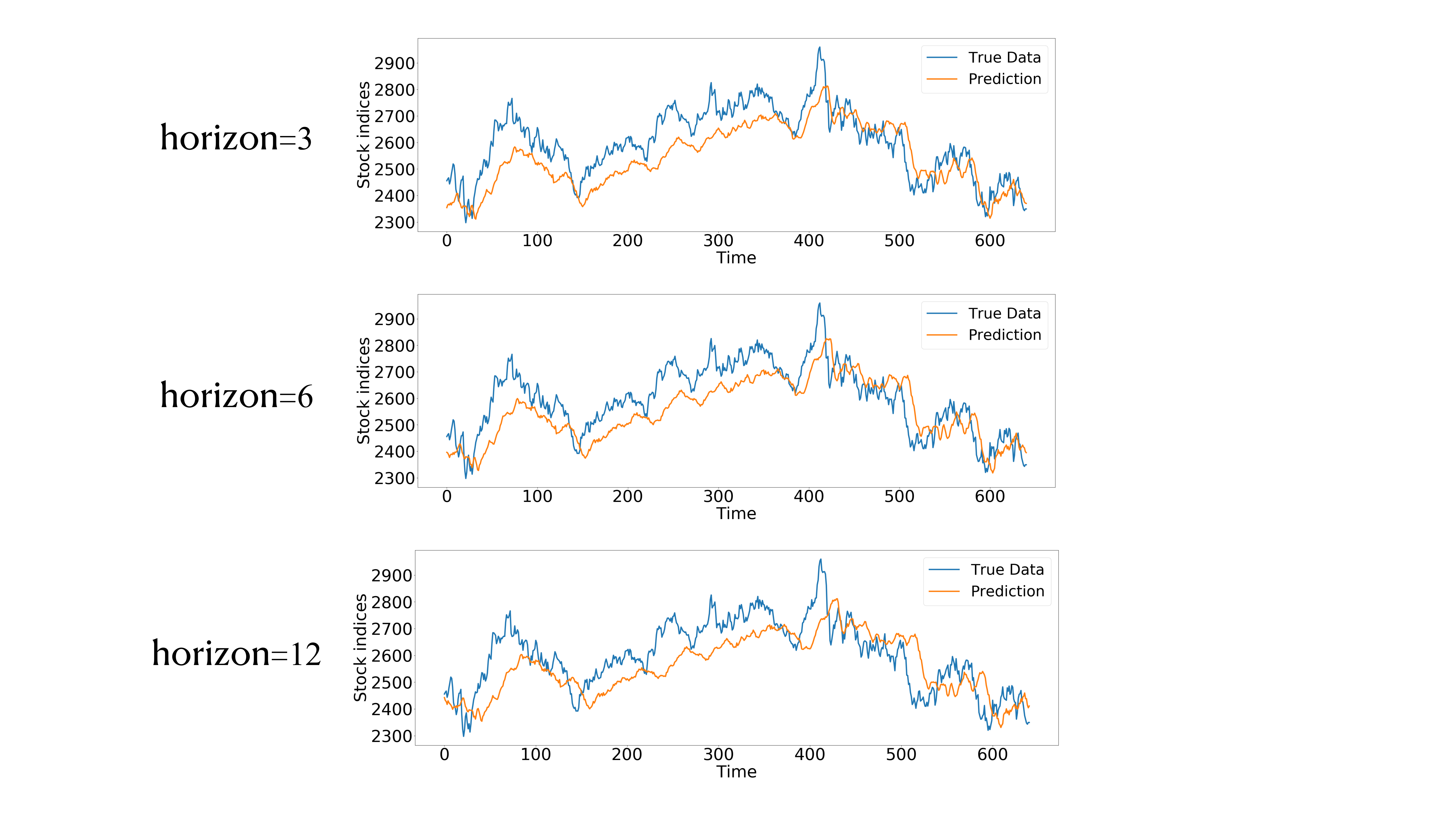}
  \caption{Prediction results for self-attention encoder in series saliency on industry stock indicies with  horizon=$3,6,12$ and window size=64. The feature is No.1 stock indicies of Hangseng Stock Composite Index.}
  \label{fig:indust} 
\end{figure*}

Because the self attention encoder obtains most of the best performance result, we evaluate the forecasting results of self attention encoder visually in series saliency on total three datasets. As shown in Figure~\ref{fig:air},\ref{fig:ele},\ref{fig:indust}, our method gives accurate forecasts. Proposed model clearly yields better forecasts around the flat line after the peak and in the valley.

\begin{figure*}[htbp] 
  \centering 
  \includegraphics[scale=0.2]{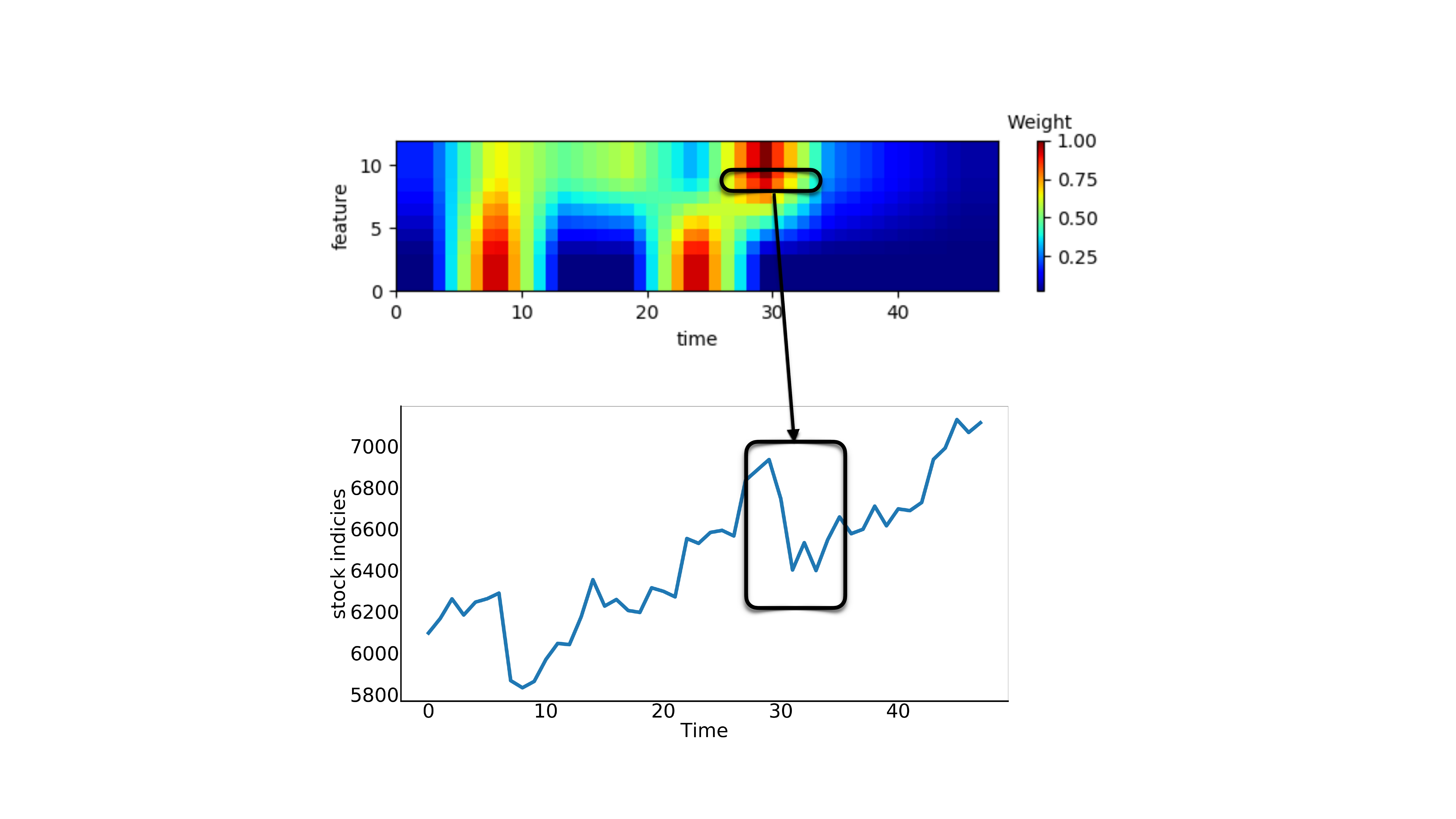}
  \caption{Series saliency for self-attention encoder  on industry stock indicies with  horizon=$6$ and window size=64. As shown in the Figure, the highlighted area corresponds to the dramatic change in the industry stock indices. The phenomenon shows that the stock indices influences the forecasting greatly.}
  \label{fig:inter1} 
\end{figure*}

\begin{figure*}[htbp] 
  \centering 
  \includegraphics[scale=0.2]{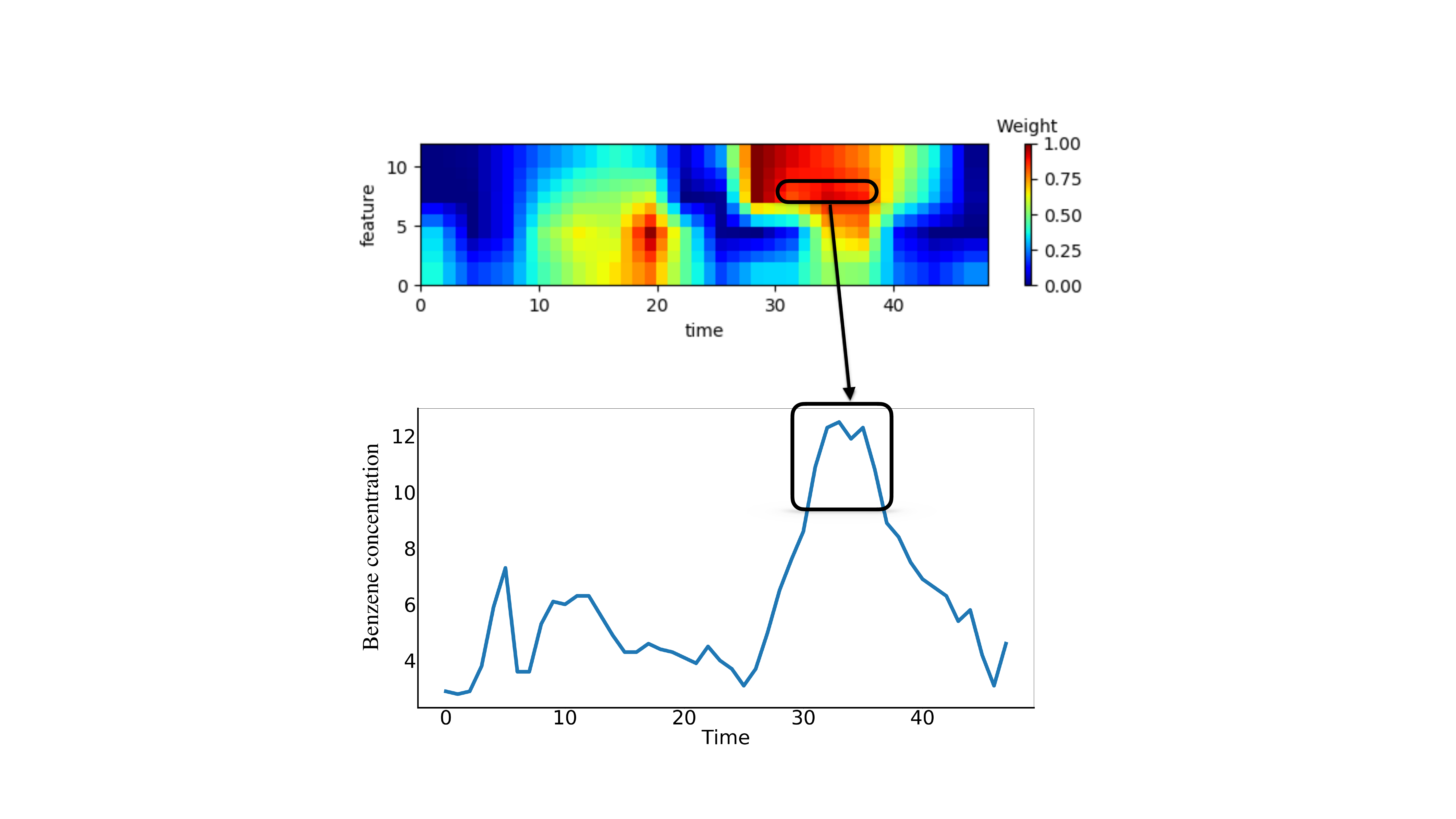}
  \caption{Series saliency for self-attention encoder on air quality with  horizon=$6$ and window size=64. The highlighted area corresponds that benzene increases sharply in air quality. As you known, benzene is harmful, which impacts air quality greatly.}
  \label{fig:inter2} 
\end{figure*}

\label{appendix:feaure}
Because the self attention encoder obtains most of the best performance result, we visualize the series saliency on the other datasets, including the air quality and industry stock composite index, for horizon=$6$. As shown in Figure \ref{fig:inter1} and Figure \ref{fig:inter2}, we could analyze the series saliency and obtain global feature importance.

\end{document}